\documentclass{article} % For LaTeX2e
\usepackage{iclr2015,times}
\usepackage{hyperref}
\usepackage{url}
\usepackage{graphicx}
\usepackage{subcaption}
\usepackage{wrapfig}
\usepackage{amsmath, amssymb}

\title{Learning Activation Functions to Improve Deep Neural Networks}

\author{
Forest Agostinelli\\
Department of Computer Science\\
University of California - Irvine\\
Irvine, CA 92697, USA \\
\texttt{\{fagostin\}@uci.edu} \\
\And
Matthew Hoffman \\
Adobe Research \\
San Francisco, CA 94103, USA\\
\texttt{\{mathoffm\}@adobe.com} \\
\And
Peter Sadowski, Pierre Baldi\\
Department of Computer Science\\
University of California - Irvine\\
Irvine, CA 92697, USA \\
\texttt{\{peter.j.sadowski,pfbaldi\}@uci.edu} \\
}

\newtheorem{theorem}{Theorem}
\newenvironment{proof}[1][Proof]{\begin{trivlist}
\item[\hskip \labelsep {\bfseries #1}]}{\end{trivlist}}

\iclrfinalcopy % Uncomment for camera-ready version

\begin{document}

\maketitle

\begin{abstract}
%The majority of neural networks use fixed activation functions that are the same for every neuron. We have designed an activation function that is learned through back-propagation. With this learned activation function, we were able to decrease error rates of baselines by up to $9.4\%$ and our results are competitive with the state of the art on image classification for Cifar10 and Cifar100; it has also made detecting the presence of the Higgs Boson more reliable.
Artificial neural networks typically have a fixed, non-linear activation function at each neuron. We have designed a novel form of piecewise linear activation function that is learned independently for each neuron using gradient descent. With this adaptive activation function, we are able to improve upon deep neural network architectures composed of static rectified linear units, achieving state-of-the-art performance on CIFAR-10 (7.51\%), CIFAR-100 (30.83\%), and a benchmark from high-energy physics involving Higgs boson decay modes.
\end{abstract}

\section{Introduction}

Deep learning with artificial neural networks has enabled rapid progress on applications in engineering \citep[e.g.,][]{krizhevsky2012imagenet, deepspeech} and basic science \citep[e.g.,][]{deepcontact2012, lusci2013deep, baldi2014searching}. Usually, the parameters in the linear components are learned to fit the data, while the nonlinearities are pre-specified to be a logistic, tanh, rectified linear, or max-pooling function.
A sufficiently large neural network using any of these common nonlinear functions can approximate arbitrarily complex functions \citep{hornik1989multilayer,cho2010arccos},
but in finite networks the choice of nonlinearity affects both the learning dynamics (especially in deep networks) and the network's expressive power.

Designing activation functions that enable fast training of accurate deep neural networks is an active area of research.
The rectified linear activation function \citep{jarrett2009best,glorot2011deep}, which does not saturate like sigmoidal functions, has made it easier to quickly train deep neural networks by alleviating the difficulties of weight-initialization and vanishing gradients. Another recent innovation is the ``maxout'' activation function, which has achieved state-of-the-art performance on multiple machine learning benchmarks~\citep{goodfellow2013maxout}. The maxout activation function computes the maximum of a set of linear functions, and has the property that it can approximate any convex function of the input. \citet{springenberg2013improving} replaced the $\max$ function with a probabilistic $\max$ function and \citet{gulcehre2014learned} explored an activation function that replaces the $\max$ function with an $L_P$ norm. However, while the type of activation function can have a significant impact on learning, the space of possible functions has hardly been explored.

One way to explore this space is to \emph{learn} the activation function during training. Previous efforts to do this have largely focused on genetic and evolutionary algorithms \citep{yao1999evolving}, which attempt to select an activation function for each neuron from a pre-defined set. Recently, \citet{turner2014neuroevolution} combined this strategy with a single scaling parameter that is learned during training.

In this paper, we propose a more powerful adaptive activation function. This parametrized, piecewise linear activation function is learned independently for each neuron using gradient descent, and can represent both convex and non-convex functions of the input. Experiments demonstrate that like other piecewise linear activation functions, this works well for training deep neural networks, and we obtain state-of-the-art performance on multiple benchmark deep learning tasks.
%The activation functions in \citet{turner2014neuroevolution} were selected from a set of predetermined activation functions and had one other parameter that scaled the input.

%\input{sec-RelatedWork}

\section{Adaptive Piecewise Linear Units}
\label{sec:learnedfunctions}

% TODO: General description of the scope of our formulation.

Here we define the adaptive piecewise linear (APL) activation unit. Our method formulates the activation function $h_i(x)$ of an APL unit $i$ as a sum of hinge-shaped functions,
\begin{equation}
h_i(x) = \max(0,x) + \sum_{s=1}^{S}a_{i}^{s}\max(0,-x+b_{i}^{s})
\label{eq:learnedActiv}
\end{equation}
The result is a piecewise linear activation function. The number of hinges, $S$, is a hyperparameter set in advance, while the variables $a^s_i$, $b^s_i$  for $i\in{1,...,S}$ are learned using standard gradient descent during training. The   $a_i^s$ variables control the slopes of the linear segments, while the $b_i^s$ variables determine the locations of the hinges.

The number of additional parameters that must be learned when using these APL units is $2SM$, where $M$ is the total number of hidden units in the network. This number is small compared to the total number of weights in typical networks.

\begin{wrapfigure}{r}{0.5\textwidth}
\center
\vspace{-0.5in}
\includegraphics[scale=0.35]{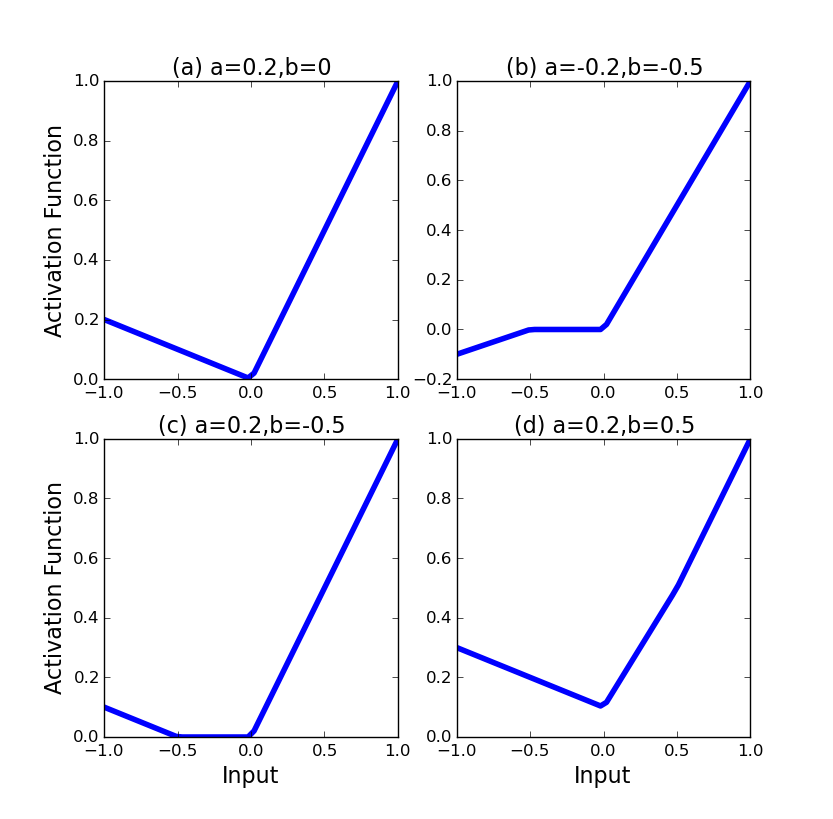}
\caption{Sample activation functions obtained from changing the parameters. Notice that figure b shows that the activation function can also be non-convex.
Asymptotically, the activation functions tend to $g(x)=x$ as $x\rightarrow\infty$ and $g(x)=\alpha x - c$ as $x\leftarrow -\infty$ for some $\alpha$ and $c$.
$S=1$ for all plots.
\vspace{-0.5in}
}
\label{fig:SampleActivFunction}
\end{wrapfigure}

Figure \ref{fig:SampleActivFunction} shows example
APL functions for $S=1$.  Note that unlike maxout, the class of functions that can be learned by a single unit includes non-convex functions.  In fact, for large
enough $S$, $h_i(x)$ can approximate arbitrarily complex continuous functions, subject to two conditions:

%% In fact, as theorem \ref{thm:universality} shows, for large enough $S$ the sum-of-hinges function can approximate arbitrary functions to arbitrary precision (over a limited domain).

%% TODO: MDH: Finish this proof to deal with the edge cases. This includes:
%% * Saying that the slope of g(x) goes to 1 as x goes to infinity
%% * Adding a hinge that doesn't go flat at -infinity.
\begin{theorem}
\label{thm:universality}
Any continuous piecewise-linear function $g(x)$ can be expressed by
Equation \ref{eq:learnedActiv} for some $S$, and $a_i$, $b_i$, $i\in{1,...,S}$, assuming that:
%% \\$\textstyle h(x)\equiv\max(0,x) + \sum_{s=1}^{S}
%% a^{s}\max(0,-x+b^{s})$ for some $S$, $a$, and $b$, assuming that:
\begin{enumerate}
\item There is a scalar $u$ such that $g(x) = x$ for all $x\ge u$.
\item There are two scalars $v$ and $\alpha$ such that $\nabla_x g(x) = \alpha$
for all $x < v$.
\end{enumerate}
\end{theorem}
This theorem implies that we can reconstruct any piecewise-linear
function $g(x)$ over any subset of the real line, and the two
conditions on $g(x)$ constrain the behavior of $g(x)$ to be linear as
$x$ gets very large or small. The first condition is less restrictive
than it may seem. In neural networks, $g(x)$ is generally only of
interest as an input to a linear function $wg(x) + z$; this linear
function effectively restores the two degrees of freedom that are
eliminated by constraining the rightmost segment of $g(x)$ to have
unit slope and bias 0.

\begin{proof}
Let $g(x)$ be piecewise linear with $K+2$ linear regions separated by ordered boundary points $b^0$, $b^1$, ...,$b^K$, and let $a^k$ be the slope of the $k$-th region. Assume also that $g(x)=x$ for all $x \ge b^K$.   We show that $g(x)$ can be expressed by the following special case of Equation \ref{eq:learnedActiv}:
\begin{equation}
\begin{split}
\label{eq:construction}
h(x) &\equiv - a^0 \max(0, -x + b^0) \\
& \quad \textstyle + \sum_{k=1}^{K} a^{k}(\max(0,-x+b^{k-1}) - \max(0, -x+b^{k})) \\
& \quad \textstyle - \max(0,-x) + \max(0,x)  + \max(0, -x+b^{K}), 
\end{split}
\end{equation}
%% \begin{eqnarray}
%% \label{eq:construction}
%% h(x) &\equiv & - a^0 \max(0, -x + b^0) \nonumber \\
%% && \textstyle + \sum_{k=1}^{K} a^{k}(\max(0,-x+b^{k-1}) - \max(0, -x+b^{k})) \nonumber \\ \nonumber
%% && \textstyle - \max(0,-x) + \max(0,x)  + \max(0, -x+b^{K}), 
%% \end{eqnarray}
The first term has slope $a^0$ in the range $(-\infty, b^0)$ and $0$ elsewhere. Each element in the summation term of Equation \ref{eq:construction} has slope $a^k$ over the range $(b^{k-1}, b^k)$ and $0$ elsewhere. The last three terms together have slope $1$ when $x\in(b^K, \infty)$ and $0$ elsewhere. Now, $g(x)$ and $h(x)$ are continuous, their slopes match almost everywhere, and it is easily verified that $h(x)=g(x)=x$ for $x \ge b^K$. Thus, we conclude that $h(x)=g(x)$ for all $x$. $\square$
\end{proof}

\subsection{Comparison with Other Activation Functions}
%% In this section we show how one could construct our learned activation
%% functions using other nonlinearities as building blocks. Specifically,
%% we will compare our learned-activation units to maxout networks
%% \citep{goodfellow2013maxout} and the network-in-network approach of
%% \citet{lin2013network}.

In this section we compare the proposed approach to learning
activation functions with two other nonlinear activation functions:
maxout \citep{goodfellow2013maxout}, and network-in-network
\citep{lin2013network}.

We observe that both maxout units and network-in-network can learn any
nonlinear activation function that APL units can, but
require many more parameters to do so. This difference allows
APL units to be applied in very different ways from
maxout and network-in-network nonlinearities: the small number of
parameters needed to tune an APL unit makes it practical
to train convolutional networks that apply different nonlinearities at
each point in each feature map, which would be completely impractical
in either maxout networks or network-in-network approaches.

\paragraph{Maxout.}
Maxout units differ from traditional neural network nonlinearities in
that they take as input the output of \emph{multiple} linear
functions, and return the largest:
\begin{equation}
\begin{split}
h_{\mathrm{maxout}}(x) = \max_{k\in\{1,\ldots,K\}}
w^k \cdot x + b^k.
\end{split}
\end{equation}
Incorporating multiple linear functions increases the expressive power
of maxout units, allowing them to approximate arbitrary convex
functions, and allowing the difference of a pair of maxout units to
approximate arbitrary functions.

%% Since pairs of maxout units can approximate arbitrary functions, a
%% maxout network could in principle 

%% Our learned-activation units can be interpreted as a special case of
%% maxout units weight-tying, just as convolutional networks can be
%% interpreted as

%% This expressive power comes at a price---the number of parameters that
%% must be learned (and the number of multiply-adds that must be
%% performed) for a maxout unit using $K$ linear functions is $K$ times
%% higher than the number of parameters associated with a traditional
%% sigmoid, tanh, Gaussian, or rectified linear unit. For example,
%% computing the simple function $|w\cdot x|$ on a $D$-dimensional input
%% using maxout units requires $2D$ parameters, half of which are negated
%% versions of the other half. Computing non-convex functions requires
%% coordination between pairs of maxout units across multiple layers of a
%% network, which requires particularly delicate tuning.

Networks of maxout units with a particular weight-tying scheme can
reproduce the output of an APL unit. The sum of terms in
Equation \ref{eq:learnedActiv} with positive coefficients (including
the initial $\max(0, x)$ term) is a convex function, and the sum of
terms with negative coefficients is a concave function. One could
approximate the convex part with one maxout unit, and the concave part
with another maxout unit:
\begin{equation}
\begin{split}
h^\mathrm{convex}(x) = \max_k c^{\mathrm{convex}}_k w\cdot x + d^{\mathrm{convex}}_k;
\quad
h^\mathrm{concave}(x) = \max_k c^{\mathrm{concave}}_k w\cdot x + d^{\mathrm{concave}}_k,
\end{split}
\end{equation}
where $c$ and $d$ are chosen so that
\begin{equation}
\begin{split}
\textstyle
h^\mathrm{convex}(x) - h^\mathrm{concave}(x)
= \max(0, w\cdot x + u) + \sum_s a^s \max(0, w\cdot x + u).
\end{split}
\end{equation}
In a standard maxout network, however, the $w$ vectors are not tied.
So implementing APL units (Equation \ref{eq:learnedActiv}) using a maxout network would require
learning $O(SK)$ times as many parameters, where $K$ is the size of
the maxout layer's input vector. Whenever the expressive power of an
APL unit is sufficient, using the more complex
maxout units is therefore a waste of computational and modeling power.

%% The large number of parameters needed to control a maxout unit make it
%% impractical in convolutional maxout networks to learn separate
%% nonlinearities for each point in a feature map. By contrast, since the
%% number of parameters controlling a learned-activation unit is
%% relatively small, we can afford to apply different nonlinearities at
%% each

%% By contrast, learned-activation units of the form in equation
%% \ref{eq:learnedActiv} can approximate arbitrarily complex functions of
%% the form $h(w\cdot x + b)$ with a minimum of additional parameters.

\paragraph{Network-in-Network.}
\citet{lin2013network} proposed replacing the simple rectified linear activation in convolutional networks with a fully
connected network whose parameters are learned from data.  This
``MLPConv'' layer couples the outputs of all filters applied to a
patch, and permits arbitrarily complex transformations of the inputs.
A depth-$M$ MLPConv layer produces an output vector $f^M_{ij}$ from an
input patch $x_{ij}$ via the series of transformations
\begin{equation}
\begin{split}
f^1_{ijk} = \max(0, w_k^1\cdot x_{ij} + b_k^1),\ldots,
f^M_{ijk} = \max(0, w_k^M\cdot f_{ij}^{M-1} + b_k^M).
\end{split}
\end{equation}
As with maxout networks, there is a weight-tying scheme that allows an
MLPConv layer to reproduce the behavior of an APL unit:
\begin{equation}
\begin{split}
\textstyle
f^1_{ijk} = \max(0, c_k w_{\kappa(k)}\cdot x_{ij} + b_{k}^1),
f^2_{ijk} = \sum_{\ell|\kappa(\ell)=k} a_k f^1_{ij\ell},
\end{split}
\end{equation}
where the function $\kappa(k)$ maps from hinge output indices $k$ to
filter indices $\kappa$, and the coefficient $c_k\in\{-1, 1\}$.

This is a very aggressive weight-tying scheme that dramatically
reduces the number of parameters used by the MLPConv layer. Again we
see that it is a waste of computational and modeling power to use
network-in-network wherever an APL unit would suffice.

%% Since fully connected networks with rectified linear units are
%% universal function approximators, one could again consider the
%% learned-activation unit of equation \ref{eq:learnedActiv} a special
%% case of the network-in-network approach. However, like maxout, the
%% network-in-network approach uses many more parameters than a
%% learned-activation unit does, and uses them to couple multiple filters
%% applied to the same patch, which the learned-activation unit cannot
%% do.

However, network-in-network can do things that APL units cannot---in
particular, it efficiently couples and summarizes the outputs of
multiple filters. One can get the benefits of both architectures by
replacing the rectified linear units in the MLPconv layer with
APL units.

%% We evaluate this approach
%% in section \ref{sec:experiments}, and find that it yields a
%% significant improvement over standard network-in-network on the
%% CIFAR-100 dataset. Also, on CIFAR-100 using learned activations in a
%% standard convolutional network outperformed network-in-network
%% (without learned activations), suggesting that one does not always
%% need the full expressive power of the network-in-network nonlinearity
%% to achieve excellent results.

\section{Experiments}
\label{sec:experiments}
Experiments were performed using the software package CAFFE \citep{jia2014caffe}. The hyperparameter, $S$, that controls the complexity of the activation function was determined using a validation set for each dataset. The $a^s_i$ and $b^s_i$ parameters were regularized with an L2 penalty, scaled by $0.001$. Without this penalty, the optimizer is free to choose very large values of $a^s_i$ balanced by very small weights, which would lead to numerical instability. We found that adding this penalty improved results. The model files and solver files are available at https://github.com/ForestAgostinelli/Learned-Activation-Functions-Source/tree/master.

\subsection{CIFAR}
The CIFAR-10 and CIFAR-100 datasets \citep{cifar} are 32x32 color images that have 10 and 100 classes, respectively. They both have 50,000 training images and 10,000 test images. The images were preprocessed by subtracting the mean values of each pixel of the training set from each image. Our network for CIFAR-10 was loosely based on the network used in \citep{srivastava2014dropout}. It had 3 convolutional layers with 96, 128, and 256 filters, respectively. Each kernel size was 5x5 and was padded by 2 pixels on each side. The convolutional layers were followed by a max-pooling, average-pooling, and average-pooling layer, respectively; all with a kernel size of 3 and a stride of 2. The two fully connected layers had 2048 units each. We applied dropout \citep{hinton2012improving} to the network as well. We found that applying dropout both before \textit{and} after a pooling layer increased classification accuracy. The probability of a unit being dropped before a pooling layer was 0.25 for all pooling layers. The probability for them being dropped after each pooling layers was 0.25, 0.25, and 0.5, respectively. The probability of a unit being dropped for the fully connected layers was 0.5 for both layers. The final layer was a softmax classification layer. For CIFAR-100, the only difference was the second pooling layer was max-pooling instead of average-pooling. The baseline used rectified linear activation functions.

When using the APL units, for CIFAR-10, we set $S=5$. For CIFAR-100 we set $S=2$. Table \ref{cifar_table} shows that adding the APL units improved the baseline by over 1\% in the case of CIFAR-10 and by almost 3\% in the case of CIFAR-100. \textit{In terms of relative difference, this is a 9.4\% and a 7.5\% decrease in error rate, respectively.} We also try the network-in-network architecture for CIFAR-10 \citep{lin2013network}. We have $S=2$ for CIFAR-10 and $S=1$ for CIFAR-100. We see that it improves performance for both datasets.

We also try our method with the augmented version of CIFAR-10 and CIFAR-100. We pad the image all around with a four pixel border of zeros. For training, we take random $32$ x $32$ crops of the image and randomly do horizontal flips. For testing we just take the center $32$ x $32$ image. To the best of our knowledge, the results we report for data augmentation using the network-in-network architecture \textit{are the best results reported for CIFAR-10 and CIFAR-100 for any method.}

In section \ref{sec:visualize}, one can observe that the learned activations can look similar to leaky rectified linear units (Leaky ReLU) \citep{maas2013rectifier}. This activation function is slightly different than the ReLU because it has a small slope $k$ when the input $x<0$.
\[
    h(x)= 
\begin{cases}
    x, & \text{if } x > 0\\
    kx,              & \text{otherwise}
\end{cases}
\]

In \citep{maas2013rectifier}, $k$ is equal to $0.01$. To compare Leaky ReLUs to our method, we try different values for $k$ and pick the best value one. The possible values are positive and negative $0.01$, $0.05$, $0.1$, and $0.2$. For the standard convolutional neural network architecture $k=0.05$ for CIFAR-10 and $k=-0.05$ for CIFAR-100. For the network-in-network architecture $k=0.05$ for CIFAR-10 and $k=0.2$ for CIFAR-100.
APL units consistently outperform leaky ReLU units, showing the value of tuning the nonlinearity (see also section \ref{sec:hyperparams}).

\begin{table}[h]
\caption{Error rates on CIFAR-10 and CIFAR-100 with and without data augmentation. This includes standard convolutional neural networks (CNNs) and the network-in-network (NIN) architecture \citep{lin2013network}. The networks were trained $5$ times using different random initializations --- we report the mean followed by the standard deviation in parenthesis. The best results are in bold.}
\label{cifar_table}
\begin{center}
\begin{tabular}{p{8.4cm}p{2.2cm}p{2.2cm}}
\bf Method  &\bf CIFAR-10 & \bf CIFAR-100\\
\hline
\end{tabular}
Without Data Augmentation
\begin{tabular}{p{8.4cm}p{2.2cm}p{2.2cm}}
\hline
CNN + ReLU \citep{srivastava2014dropout} & 12.61\% & 37.20\%\\
CNN + Channel-Out \citep{wang2013maxout} & 13.2\% & 36.59\% \\
CNN + Maxout \citep{goodfellow2013maxout} & 11.68\% & 38.57\% \\
CNN + Probout \citep{springenberg2013improving} & \bf11.35\% & 38.14\% \\
CNN (Ours) + ReLU & 12.56 (0.26)\%  & 37.34 (0.28)\%  \\
CNN (Ours) + Leaky ReLU & 11.86 (0.04)\% & 35.82 (0.34)\% \\
CNN (Ours) + APL units  & 11.38 (0.09)\% &\bf34.54 (0.19)\%  \\
\hline
NIN + ReLU \citep{lin2013network} & 10.41\% & 35.68\% \\
NIN + ReLU + Deep Supervision \citep{lee2014deeply} & 9.78\% & 34.57\% \\
NIN (Ours) + ReLU  & 9.67 (0.11)\%   &  35.96 (0.13)\% \\
NIN (Ours) + Leaky ReLU &  9.75 (0.22)\%  &  36.00 (0.36)\%  \\
NIN (Ours) + APL units & \bf9.59 (0.24)\% &  \bf34.40 (0.16)\%   \\
\hline
\end{tabular}
With Data Augmentation
\begin{tabular}{p{8.4cm}p{2.2cm}p{2.2cm}}
\hline
CNN + Maxout \citep{goodfellow2013maxout} & 9.38\% & - \\
CNN + Probout \citep{springenberg2013improving} & 9.39\% & - \\
CNN + Maxout \citep{stollenga2014deep} & 9.61\% & 34.54\% \\
CNN + Maxout + Selective Attention \citep{stollenga2014deep} & \bf9.22\% & \bf33.78\% \\
CNN (Ours) + ReLU  &  9.99 (0.09)\%  & 34.50 (0.12)\%  \\
CNN (Ours) + APL units  &  9.89 (0.19)\%  &33.88 (0.45)\%  \\
\hline
NIN + ReLU \citep{lin2013network} & 8.81\% & - \\
NIN + ReLU + Deep Supervision \citep{lee2014deeply} & 8.22\% & - \\
NIN (Ours) + ReLU & 7.73 (0.13)\%   & 32.75 (0.13)\%  \\
NIN (Ours) + APL units & \bf7.51 (0.14)\%  &  \bf30.83 (0.24)\%   \\
\hline
\end{tabular}
\end{center}
\end{table}

\subsection{Higgs Boson Decay}
The Higgs-to-$\tau^+\tau^-$ decay dataset comes from the field of high-energy physics and the analysis of data generated by the Large Hadron Collider~\citep{baldi2014enhanced}. The dataset contains 80 million collision events, characterized by 25 real-valued features describing the 3D momenta and energies of the collision products. The supervised learning task is to distinguish between two types of physical processes: one in which a Higgs boson decays into $\tau^+\tau^-$ leptons and a background process that produces a similar measurement distribution. Performance is measured in terms of the area under the receiver operating characteristic curve (AUC) on a test set of 10 million examples, and in terms of discovery significance \citep{Cowan2010js} in units of Gaussian $\sigma$, using 100 signal events and 5000 background events with a 5\% relative uncertainty.
 
Our baseline for this experiment is the 8 layer neural network architecture from \citep{baldi2014enhanced} whose architecture and training hyperparameters were optimized using the Spearmint algorithm~\citep{snoek2012practical}. We used the same architecture and training parameters except that dropout was used in the top two hidden layers to reduce overfitting. For the APL units we used $S=2$. Table \ref{tab:higgsboson} shows that a single network with APL units achieves state-of-the-art performance, increasing performance over the dropout-trained baseline and the ensemble of 5 neural networks from \citep{baldi2014enhanced}. 

\begin{table}[h]
\caption{Performance on the Higgs boson decay dataset in terms of both AUC and expected discovery significance. The networks were trained $4$ times using different random initializations --- we report the mean followed by the standard deviation in parenthesis. The best results are in bold.}
\label{tab:higgsboson}
\begin{center}
\begin{tabular}{lll}
\multicolumn{1}{c}{\bf Method}  &\multicolumn{1}{c}{\bf AUC} &\multicolumn{1}{c}{\bf Discovery Significance}\\
\hline
DNN + ReLU \citep{baldi2014enhanced} & 0.802 & 3.37$\sigma$ \\
DNN + ReLU + Ensemble\citep{baldi2014enhanced} & 0.803 & 3.39$\sigma$ \\
DNN (Ours) + ReLU & 0.803 (0.0001) & 3.38 (0.008) $\sigma$  \\
DNN (Ours) + APL units & \bf0.804 (0.0002) & \bf3.41 (0.006) $\sigma$   \\
\hline
\end{tabular}
\end{center}
\end{table}

\subsection{Effects of APL unit Hyperparameters}
\label{sec:hyperparams}
Table \ref{inputcoeff} shows the effect of varying $S$ on the CIFAR-10 benchmark. We also tested whether learning the activation function was important (as opposed to having complicated, \emph{fixed} activation functions). For $S=1$, we tried freezing the activation functions at their random initialized positions, and not allowing them to learn. The results show that learning activations, as opposed to keeping them fixed, results in better performance.

\begin{table}[h]
\caption{Classification accuracy on CIFAR-10 for varying values of $S$. Shown are the mean and standard deviation over $5$ trials.}
\label{inputcoeff}
\begin{center}
\begin{tabular}{ll}
\multicolumn{1}{c}{\bf Values of $S$}  &\multicolumn{1}{c}{\bf Error Rate}
\\ \hline
 baseline & 12.56 (0.26)\%\\
 \hline
 $S=1$ (activation not learned) & 12.55 (0.11)\%\\
 $S=1$ & 11.59 (0.16)\%\\
 \hline
 $S=2$ & 11.73 (0.23)\%\\
 \hline
 $S=5$ & \bf11.38 (0.09)\%\\
\hline
 $S=10$ & 11.60 (0.16)\%\\
 \hline
\end{tabular}
\end{center}
\end{table}

\subsection{Visualization and Analysis of Adaptive Piecewise Linear Functions}
\label{sec:visualize}
\begin{figure}[h]
\center
\includegraphics[scale=0.3]{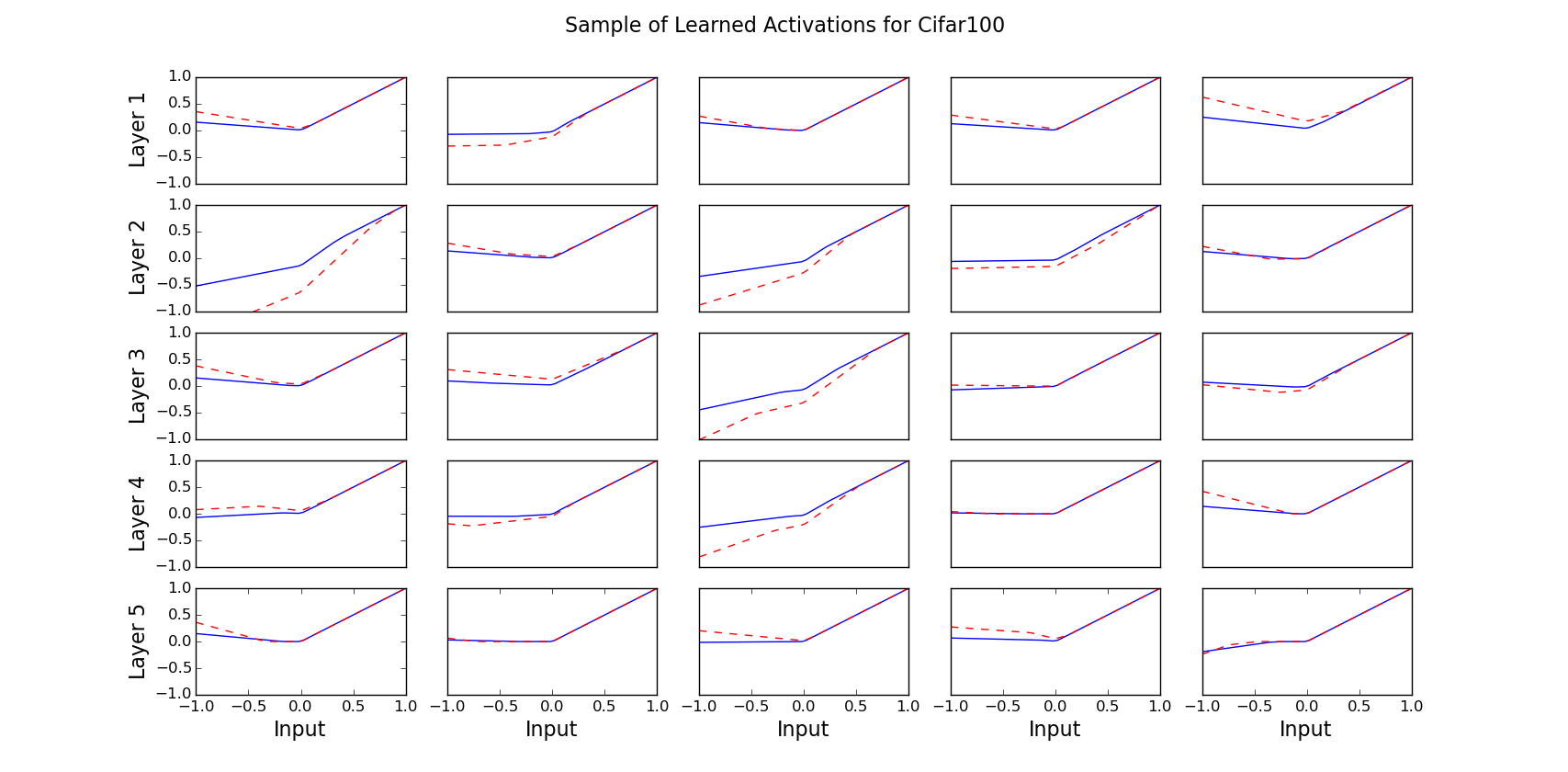}
\caption{CIFAR-100 Sample Activation Functions. Initialization (dashed line) and the final learned function (solid line).}
\label{fig:Cifar100SampleActivFunction}
\end{figure}

\begin{figure}[h]
\center
\includegraphics[scale=0.3]{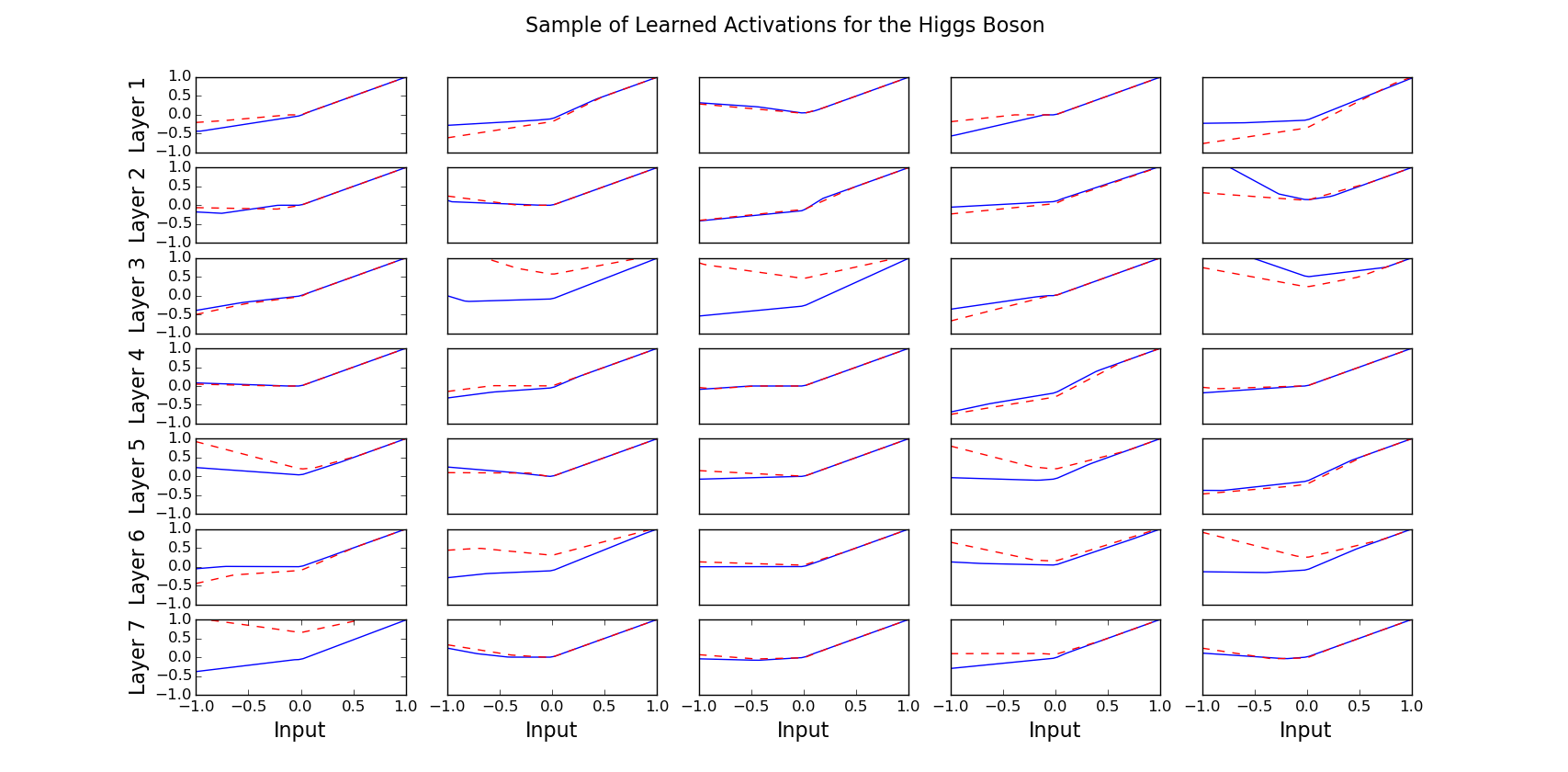}
\caption{Higgs$\rightarrow\tau^+\tau^-$ Sample Activation Functions. Initialization (dashed line) and the final learned function (solid line).}
\label{fig:HiggsBosonSampleActivFunction}
\end{figure}

The diversity of adaptive piecewise linear functions was visualized by plotting $h_i(x)$ for sample neurons. Figures \ref{fig:Cifar100SampleActivFunction} and \ref{fig:HiggsBosonSampleActivFunction} show adaptive piecewise linear functions for the CIFAR-100 and Higgs$\rightarrow\tau^+\tau^-$ experiments, along with the random initialization of that function.

\begin{figure}[h]
\begin{subfigure}[h]{0.5\textwidth}
\center
\includegraphics[scale=0.3]{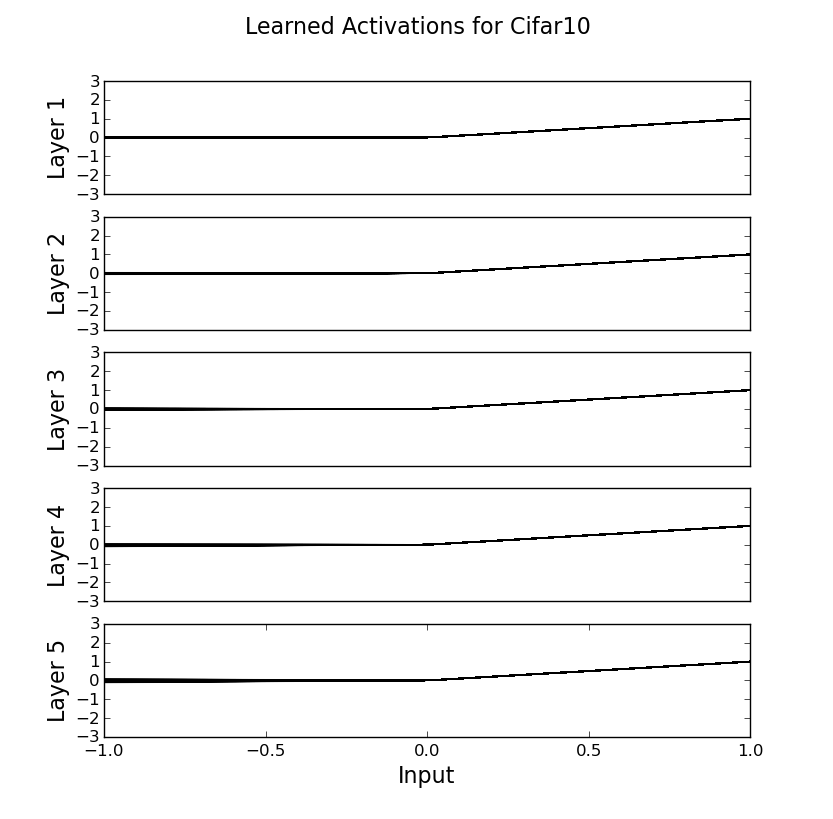}
\caption{CIFAR-10 Activation Functions.}
\label{fig:Cifar10ActivFunction}
\end{subfigure}
\begin{subfigure}[h]{0.5\textwidth}
\center
\includegraphics[scale=0.3]{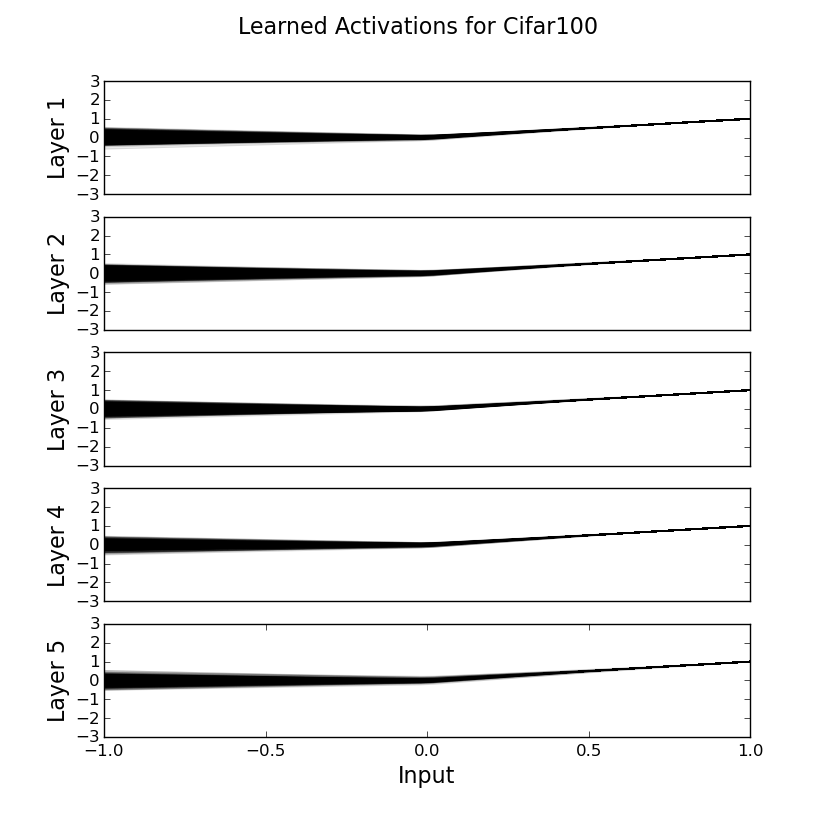}
\caption{CIFAR-100 Activation Functions.}
\label{fig:Cifar100ActivFunction}
\end{subfigure}
\begin{subfigure}[h]{0.5\textwidth}
\center
\includegraphics[scale=0.3]{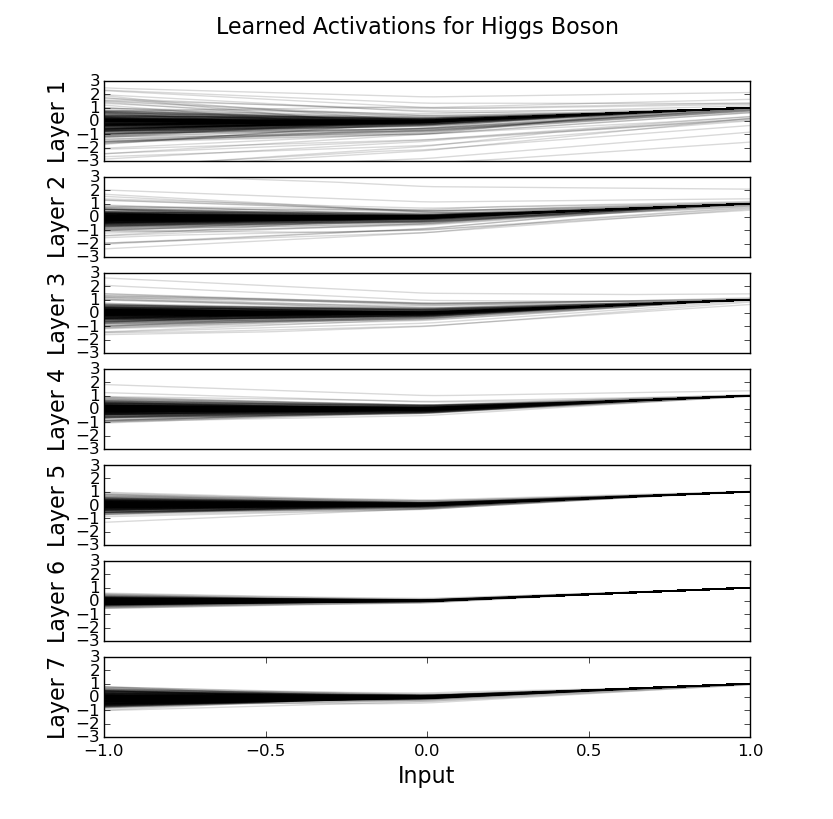}
\caption{Higgs$\rightarrow\tau^+\tau^-$ Activation Functions.}
\label{fig:HiggsBosonActivFunction}
\end{subfigure}
\caption{Visualization of the range of the values for the learned activation functions for the deep neural network for the CIFAR datasets and Higgs$\rightarrow\tau^+\tau^-$ dataset.}
\label{fig:activfunction}
\end{figure}

In figure \ref{fig:activfunction}, for each layer, 1000 activation functions (or the maximum number of activation functions for that layer, whichever is smaller) are plotted. One can see that there is greater variance in the learned activations for CIFAR-100 than there is for CIFAR-10. There is greater variance in the learned activations for Higgs$\rightarrow\tau^+\tau^-$ than there is for CIFAR-100. For the case of Higgs$\rightarrow\tau^+\tau^-$, a trend that can be seen is that the variance decreases in the higher layers.

%The activation functions differ mostly in the domain $x<0$, so next we group functions for which $h_i(x=-1) < 0$ and those for which $h_i(x=-1) > 0$. Figures \ref{fig:Cifar10ActivFunction}, \ref{fig:Cifar100ActivFunction}, and \ref{fig:HiggsBosonActivFunction} plot the average value of each set of learned functions, with $h_i(x=-1) < 0$ shown in green and $h_i(x=-1) > 0$ shown in blue. The plots also show the standard deviation, and how many neurons fall into each category. The activation functions have an equal chance of falling into each category during the random initialization.

%The learned activations for CIFAR-10 are similar to the rectifier function, with the activations very close to zero for $x<0$. However, none of the slopes are exactly zero, and this slight difference leads to a $9.4$\% decrease in error rate. We can see that in the fully connected layers, layers 4 and 5, there are far more neurons that fall into the category of $h_i(x=-1) < 0$. This also holds true for the average-pooling layers, layers 2 and 3, though to a lesser extent. This behavior is not observed in the CIFAR-100 dataset shown in Figure \ref{fig:Cifar100ActivFunction}.

%In the Higgs$\rightarrow\tau^+ \tau^-$ experiment (Figure \ref{fig:HiggsBosonActivFunction}), the activation functions have greater variation than in the CIFAR-10 and CIFAR-100 datasets. The graph also reveals a pattern in which both the mean and standard deviation decrease in the higher layers.

%Since the activations for some datasets can be very similar to a rectified linear activation, we would like to see if our method simply assists with training or actually has a significant impact on the network's output after training. To do this, we plug in the values for the weights and biases for the convolutional and fully connected layers into an identical network that has a rectified linear activation instead of the learned activations to see how it affects accuracy. The results in table \ref{tab:replacelearn} show that the learned activation functions are crucial to the performance of the network.

%\begin{table}[h]
%\caption{Performance for a network trained with APL units and that same network whose APL units are replaced with rectified linear units (ReLUs)}
%\label{tab:replacelearn}
%\begin{center}
%\begin{tabular}{llll}
%\multicolumn{1}{c}{\bf Method}  &\multicolumn{1}{c}{\bf CIFAR-10} &\multicolumn{1}{c}{\bf CIFAR-100} &\multicolumn{1}{c}{\bf $Higgs\rightarrow\tau^+\tau^-$}\\
%\hline
%Network trained with APL units & 11.45\% & 34.22\% & 0.804/3.40$\sigma$ \\
%APL units replaced with ReLUs &  50.75\% & 48.1\% & 0.544/0.514$\sigma$\\
%\hline
%\end{tabular}
%\end{center}
%\end{table}

\section{Conclusion}
We have introduced a novel neural network activation function in which each neuron computes an independent, piecewise linear function. The parameters of each neuron-specific activation function are learned via gradient descent along with the network's weight parameters. Our experiments demonstrate that learning the activation functions in this way can lead to significant performance improvements in deep neural networks without significantly increasing the number of parameters. Furthermore, the networks learn a diverse set of activation functions, suggesting that the standard one-activation-function-fits-all approach may be suboptimal. 

\subsubsection*{Acknowledgments}
F. Agostinelli was supported by the GEM Fellowship. This work was done during an internship at Adobe. We also wish to acknowledge the support of NVIDIA Corporation with the donation of the Tesla K40 GPU used for this research, NSF grant IIS-0513376, and a Google Faculty Research award to P. Baldi, and thanks to Yuzo Kanomata for computing support.

\bibliography{sec-bib}
\bibliographystyle{iclr2015}

\end{document}